\documentclass[10pt,twocolumn,letterpaper]{article}

\usepackage{cvpr}              
\definecolor{cvprblue}{rgb}{0.21,0.49,0.74}
\usepackage[pagebackref,breaklinks,colorlinks,allcolors=cvprblue]{hyperref}
\usepackage{algorithm}
\usepackage{algpseudocode}
\usepackage{amsmath}
\usepackage{makecell}
\usepackage{tabularx}
\usepackage{bbding}
\usepackage{amssymb}
\usepackage{xcolor}
\usepackage{colortbl}
\usepackage{graphicx}
\usepackage{adjustbox}
\usepackage{capt-of}
\usepackage{multirow}
\usepackage{multicol}
\usepackage{booktabs}
\usepackage{arydshln}
\usepackage{cuted}
\usepackage{tgadventor}
\usepackage{textcomp}
\usepackage[misc]{ifsym}
\definecolor{grey}{RGB}{240,240,240}
\definecolor{blue}{RGB}{217,232,242}
\definecolor{green}{RGB}{16,173,81}
\definecolor{red}{RGB}{175,37,25}
\def\paperID{36272} 
\def\confName{CVPR}
\def\confYear{2026}

\title{Score2Instruct: Scaling Up Video Quality-Centric Instructions \\via Automated Dimension Scoring}

\author{Qizhi Xie$^{1,2}$, ~Kun Yuan$^{2\textrm{ \Letter}}$, ~Yunpeng Qu$^{1,2}$, ~Jiachao Gong$^{2}$, ~Mingda Wu$^{2}$, \\ Ming Sun$^{2}$, Chao Zhou$^{2}$, Jihong Zhu$^{1\textrm{ \Letter}}$ \\
\textsuperscript{\rm 1} Tsinghua University,  \textsuperscript{\rm 2}Kuaishou Technology \\
{\tt \small xqz20@mail.tsinghua.edu.cn,}
{\tt \small yuankun03@kuaishou.com,}
{\tt \small jhzhu@tsinghua.edu.cn}
}
\begin{document}
\maketitle
\begin{strip}
    \centering
    \includegraphics[width=1.0\linewidth]{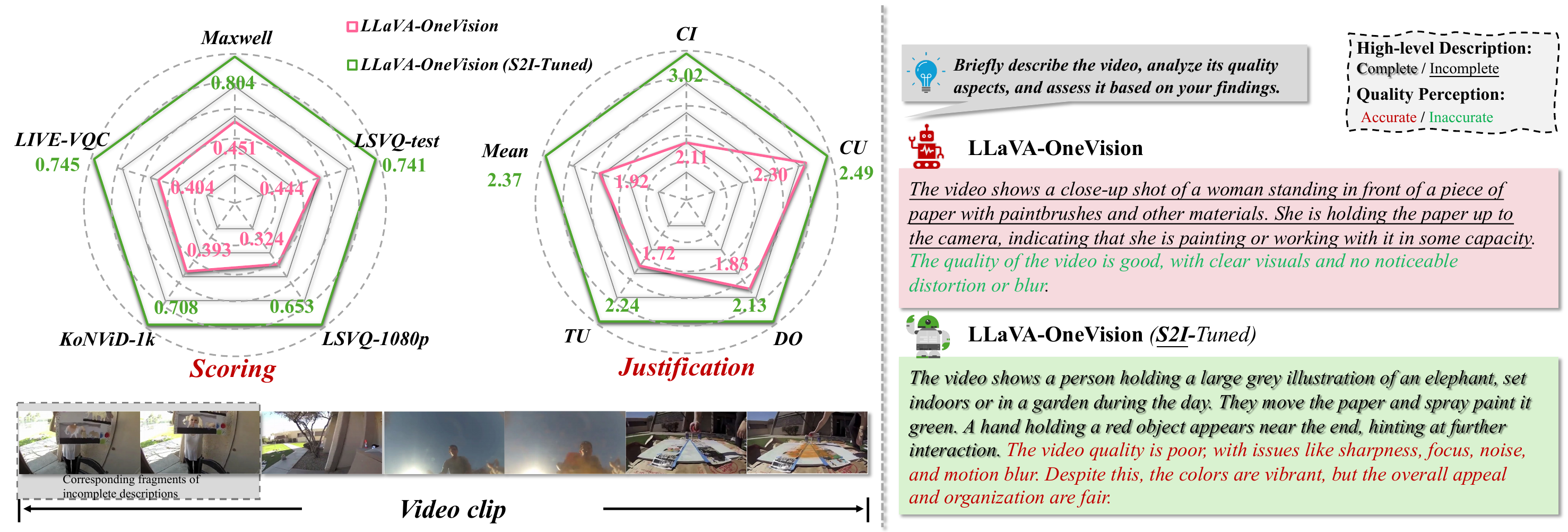}
    \captionof{figure}{
    Performance gains across video quality assessment and justification using Score2Instruct, evaluated by SRCC and VCG scores \cite{videochatgpt} respectively (\textit{Left}). An example from a video clip is also presented, demonstrating that the model fine-tuned on the proposed Score2Instruct dataset produces precise and comprehensive quality justifications (\textit{Right}).
    }
    \vspace{-6pt}
    \label{fig:teaser}
\end{strip}

\begin{abstract}
Classical video quality assessment (VQA) methods generate a numerical score to judge a video's perceived visual fidelity and clarity. 
Yet, a score fails to describe the video's complex quality dimensions (\eg, noise), restricting its applicability.
Benefiting from the human-friendly linguistic output, adapting video large multimodal models (LMMs) to VQA via instruction tuning has the potential to address this issue.
The core of the approach lies in the video quality-centric instruction data. 
Previous explorations mainly focus on the image domain, and their data generation processes heavily rely on human quality annotations and proprietary systems (\eg, GPT-4), limiting data scalability and effectiveness.
To address these challenges, we propose the \textbf{S}core-based \textbf{I}nstruction \textbf{G}eneration (\textbf{SIG}) pipeline.
Specifically, SIG first scores multiple quality dimensions of an unlabeled video and maps scores to text-defined levels. 
It then explicitly incorporates a hierarchical Chain-of-Thought (CoT) to model the correlation between specific dimensions and overall quality, mimicking the human visual system's (HVS) reasoning process.
The automated pipeline eliminates the reliance on expert-written quality descriptions and proprietary systems, ensuring data scalability and generation efficiency.
To this end, the resulting \textbf{S}core\textbf{2I}nstruct (\textbf{S2I}) dataset contains over 320K diverse instruction-response pairs, laying the basis for instruction tuning. 
Moreover, to advance video LMMs' quality scoring and justification abilities simultaneously, we devise a progressive tuning strategy to unleash the power of S2I fully.
Built upon SIG, we further curate a benchmark termed \textbf{S2I-Bench} with 400 open-ended questions to better evaluate the quality justification capacity of video LMMs.
Experimental results on the S2I-Bench and existing benchmarks indicate that our method consistently improves quality scoring and justification capabilities across multiple video LMMs.
The code and dataset will be available at \url{https://github.com/KeiChiTse/S2I}.
\end{abstract}
\vspace{-3mm}

\section{Introduction}
\label{intro}

Video quality assessment (VQA) aims to automatically evaluate the perceptual quality of input videos, imitating humans' subjective feedback when viewing a video.
Since perceptual quality greatly impacts the Quality of Experience (QoE), VQA has been extensively studied and has a variety of applications such as video enhancement, transcoding, and transmission \cite{compress,sr,xpsr,img_inpaint}. 
Classical VQA methods \cite{clipiqa,hvs1,hvs2} output a \textit{numerical score} to represent the overall quality. 
Despite the remarkable progress driven by deep learning, these methods continue to exhibit a substantial gap in VQA compared to humans since \textit{an overall quality score is insufficient to describe the complex and interrelated quality dimensions} (\eg, flicker, motion blur \etc) within the videos \cite{qbench,depictqa,mllmiqa}, limiting their practical usage.
Moreover, with the surge of User-Generated Content (UGC) \cite{kvq,vqt} and AI-generated content (AIGC) \cite{aigc_survey2,aigc_survey3} videos recently, the need to \textit{interpret video quality more versatilely} has become increasingly pronounced. 

Luckily, the advent of large multimodal models (LMMs) \cite{llava,blip2,llava1.5,internvl,qwenvl} has created new possibilities for addressing this demand. 
Based on the language decoder, it becomes feasible for models to interact with humans seamlessly through natural language with \textbf{quality justifications} \cite{qinstruct,clipiqa}, which \textit{describe video quality, reason factors leading to overall quality, and give solutions for quality improvement if possible} (Fig.\ref{fig:teaser}). 
Motivated by this, early explorations \cite{qbench,qbenchvideo,aesbench} have shown that pretrained LMMs exhibit preliminary and relatively imprecise quality assessment skill. 
One promising methodology to further augment this capability is visual instruction tuning \cite{llava,llava1.5,instructblip,rlhf}. Thus, several follow-up efforts \cite{qinstruct,coinstruct,depictqa,depictqav2,visualcritic,uniaa,aesexpert,qinstructvideo,qalign,deqa} are made by \textit{centering on generating quality-centric instructions}.
Appealing at first glance, these studies still suffer from several limitations. 
\textbf{First}, the instruction generation heavily relies on human subjective studies and proprietary APIs. 
For instance, Q-Instruct \cite{qinstruct} invites 39 experts to write 58K detailed quality descriptions (46.4 words on average) for 18,973 images and expand feedback to various types of instructions via Chat-GPT \cite{gpt3.5turbo}. 
Yet, human annotations are inherently time-consuming and susceptible to bias, largely constraining the data scaling and generation efficiency \cite{mllmiqa,ptmvqa,qpt,qptv2}. 
\textbf{Then}, previous research is mostly conducted in image quality/aesthetics assessment (IQA/IAA), lacking a deeper understanding of temporal-related factors \cite{qbenchvideo}. 
The complexities in video quality make it more challenging for humans to write in-depth quality labels from scratch than images. 
\textbf{Last}, the work above fails to enable a model to possess advanced quality scoring and justification capabilities \textit{simultaneously}, hindering the development of comprehensive quality assessors.

\begin{figure*}[t]
  \centering
  \includegraphics[width=0.9\linewidth]{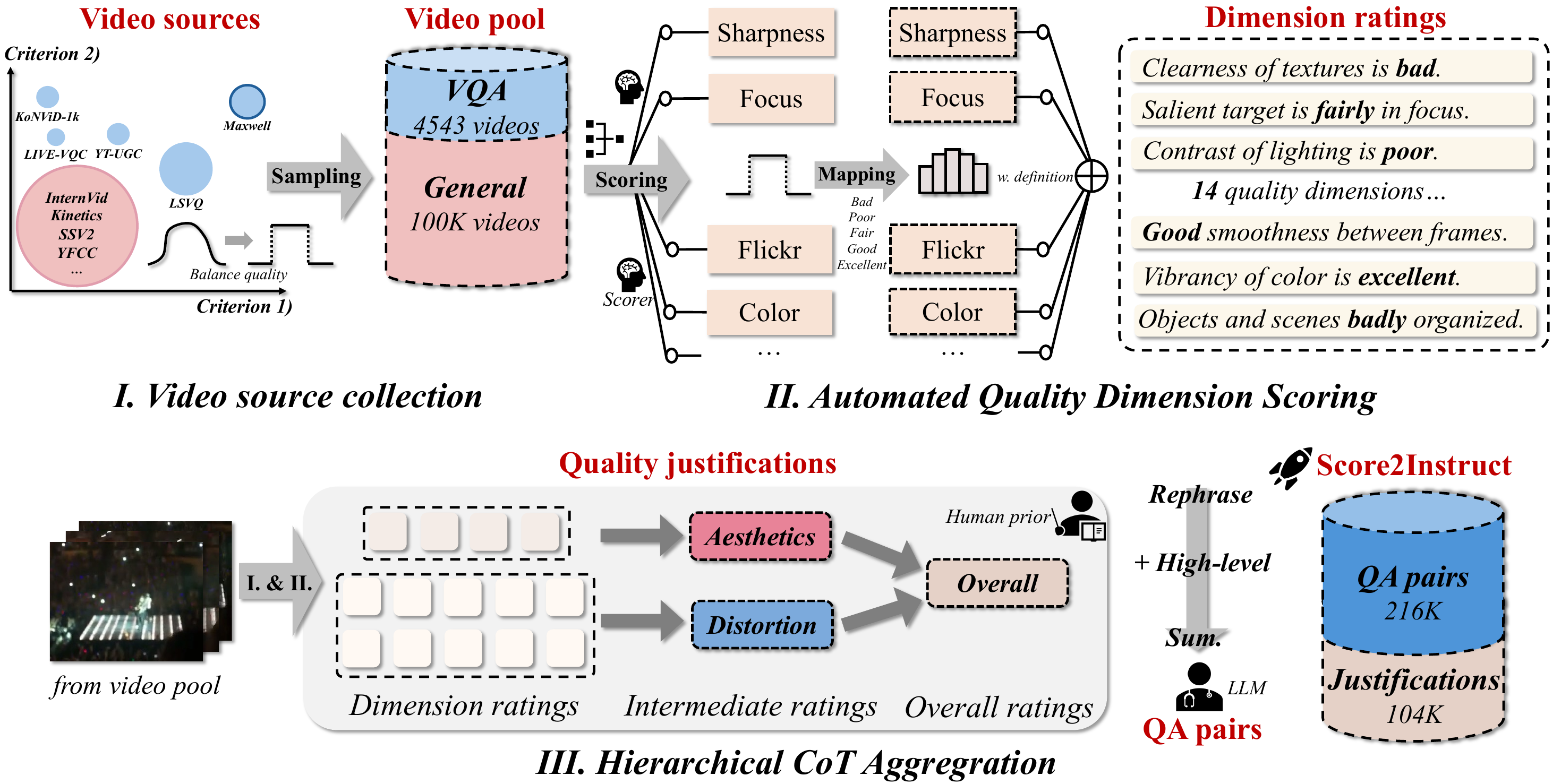}
  \caption{
  Overview of the Score-based Instruction Generation. It first samples more than 100K videos from VQA and general databases based on specific criteria. It then evaluates 14 quality dimensions to produce detailed dimension-wise ratings. Finally, a hierarchical chain-of-thought is applied to these ratings to derive the full justifications, while an LLM further expands the dataset with additional QA pairs.}
  \label{fig:framework}
\end{figure*}

In this work, we present the \textbf{S}core-based \textbf{I}nstruction \textbf{G}eneration (\textbf{SIG}) pipeline that addresses the challenges.
As in Fig.\ref{fig:framework}, SIG dissects the generation process into three key steps: \textit{video source collection}, \textit{automated quality dimension scoring}, and \textit{hierarchical CoT aggregation}.
\textbf{First}, SIG breaks the size constraint of small-scale VQA datasets by sampling videos from databases in other vision tasks. This step collects over 100K videos with balanced quality distribution via predefined criteria.
\textbf{Then}, SIG scores 14 distinct quality dimensions automatically and maps them to text-defined levels \cite{itu}, covering video-specific quality issues for subsequent model tuning.
\textbf{Last}, SIG simulates the quality reasoning process of HVS by designing a hierarchical CoT to aggregate quality dimension levels into overall justifications. Benefiting from the scalability and efficiency of SIG, an auto-generated video quality-centric instruction dataset named \textbf{S}core\textbf{2I}nstruct (\textbf{S2I}) is constructed without the help of expert annotations and proprietary APIs.

\begin{figure}[t]
  \centering
  \includegraphics[width=\linewidth]{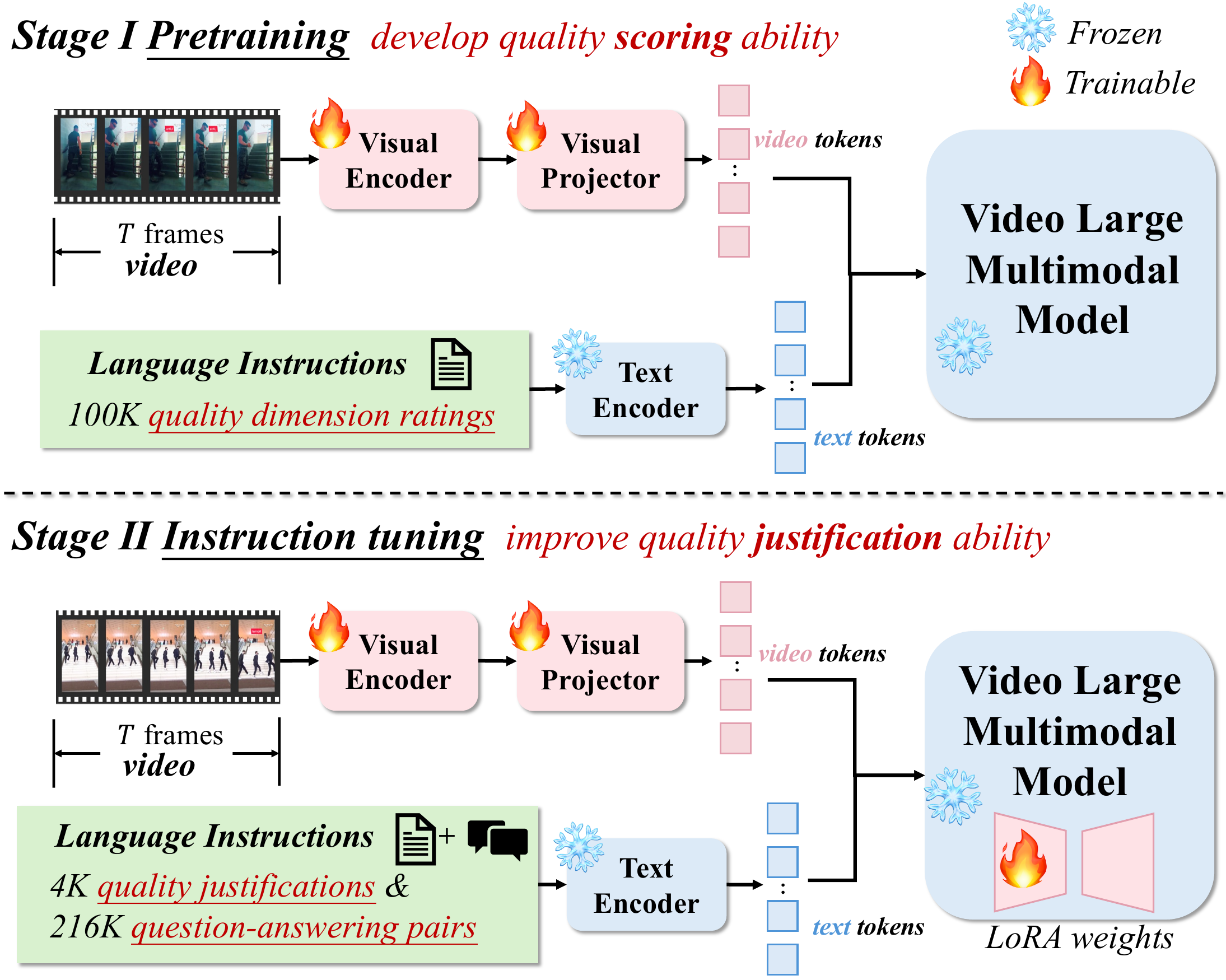}
  \caption{
  Illustration of the progressive tuning. 
  In the first stage, the model is trained on coarsely annotated data to acquire an initial sense of quality. In the second stage, it is further trained on higher-quality and more diverse data, enhancing both its scoring capability and its ability to provide justifications.
  }
  \label{fig:train}
\end{figure}

S2I contains 216K question-answering (QA) pairs and 104K justifications. 
To equip video LMMs with quality scoring and justification abilities simultaneously, we develop a two-stage tuning strategy (Fig.\ref{fig:train}) to exploit S2I rather than following the standard tuning procedures \cite{llava}. 
Moreover, we observe that current GPT-assisted benchmarks only measure the accuracy, completeness, and relevance between model responses and ground truth to assess the quality justification ability \cite{qbench,qbenchvideo,aesbench}, and the quality dimensions covered by ground truth are not comprehensive. 
Thus, we curate a benchmark dubbed \textbf{S2I-Bench} aided by SIG, which has 400 open-ended questions to facilitate a well-rounded evaluation. Our contributions are fourfold:
\begin{itemize}
    \item We introduce \textbf{SIG}, a scalable and efficient \textit{pipeline} for automated video quality-centric instruction generation. 
    \item Based on SIG, we construct \textbf{S2I}, an instruction tuning \textit{dataset} focuses on question-answering and reasoning related to VQA.
    \item Empowered by SIG, we curate \textbf{S2I-Bench}, an open-ended \textit{benchmark} that enables the thorough evaluation of quality justifications.
    \item Aided by the proposed progressive \textit{tuning strategy}, extensive experiments on six open-source and three closed-source models validate the efficacy of our methodology. Models tuned on S2I can predict precise quality scores and output reliable quality justifications at the same time.
\end{itemize}

\begin{table}[t]
\belowrulesep=0pt
\aboverulesep=0pt
\centering
\caption{Comparison of the richness of quality annotations across common VQA datasets. \# means ``the number of".}
\label{tab: vqa}
\small
\begin{tabular}{c|cc:c}
\toprule
\textbf{Database} & \textit{\#Video} & \textit{\#Score} & \textit{\#Dimension} \\ \hdashline
\textit{Youtube-UGC} & 1380 & 600K & 1 (MOS) \\
\textit{KoNViD-1k} & 1200 & 205K & 1 (MOS) \\
\textit{LIVE-VQC} & 585 & 205K & 1 (MOS) \\
\textit{LSVQ} & 39076 & 5.16M & 1 (MOS) \\ \hdashline
\textit{Maxwell} & 4543 & 2.54M & 13 \\ \bottomrule
\end{tabular}
\end{table}

\section{Related Work}
\label{related}

\noindent \textbf{Scoring-based VQA.} Classical VQA aims to accurately score the overall video quality.
At the early stage, handcrafted features based on natural scene statistics (NSS) dominate the realm of VQA \cite{nss, niqe, brisque}.
Later, deep learning-based work makes noticeable progress by directly regressing the collected Mean Opinion Scores (MOS) \cite{kvq,pvq,fastvqa,qpt,qptv2}.
One shortcoming of these methods is that they can only predict the overall quality score, lacking the capability to interpret the impact of diverse quality dimensions on it \cite{depictqa,clipiqa,maxwell}.
To solve this problem, pioneering methods \cite{liqe,clipiqa,dhqi} adapt CLIP \cite{clip} to VQA, leveraging its text encoder to input human queries for perceiving various quality attributes (\eg, blurring). 
Still, CLIP-based methods are restricted to the text encoder's fixed interface, showing limited interactivity and adaptability to the user's instructions \cite{llava}. 
Although scoring-based methods suffer from the shortcomings above, accurately scoring video quality is still of utmost importance for VQA. In this work, we take a different view to leverage the scalability of automated quality scoring in the context of video LMMs. The proposed SIG and S2I allow us to achieve precise quality scoring and justification simultaneously.

\noindent \textbf{LMMs for quality and aesthetics assessment.}
By combining the power of visual encoder (\eg, CLIP \cite{clip}, SigLIP \cite{siglip}) and large language models (LLMs) (\eg, LLaMA \cite{llama}), LMMs \cite{gpt4v, llava, internlm, blip2} exhibit remarkable high-level visual comprehension capabilities across a wide range of tasks.
In contrast, the quality and aesthetics assessment of LMMs remains less than satisfactory \cite{depictqa,qbench,qbenchvideo,aesbench}.
Thus, previous work taps into visual instruction tuning to enhance this capability.
Among them, \cite{qalign,qboost,compare2score,deqa} solely focus on improving IQA scoring accuracy.
Conversely, \cite{qinstruct,visualcritic,depictqa,depictqav2,coinstruct} improve image quality justification skill by building instruction datasets via proprietary LLMs \cite{gpt3,gpt3.5turbo} or LMMs \cite{gpt4v}. 
Similar endeavors are made in IAA \cite{aesexpert,uniaa}.
Nonetheless, these studies are restricted to the image domain. 
Inspired by \cite{qinstruct,coinstruct,qalign}, \cite{qinstructvideo} adopts a similar data generation pipeline in the video domain. Albeit with progress, its data scaling and generation efficiency are still restricted by VQA databases and manual annotations. 
In contrast, SIG exploits unlabeled videos and automatically generates large-scale video quality-centric instructions containing quality dimension descriptions with reasoning.

\section{SIG: Score-based Instruction Generation}
\label{method: sig}

The pivotal catalyst for subsequent tuning is generating a video quality-centric instruction dataset at scale.
To secure data scalability and generation efficiency, SIG features tailored designs in the aspects of video source collection, automated quality dimension scoring, and hierarchical CoT aggregation, described next.

\subsection{Video Source Collection}
\label{method: sig_video}

Previous approaches extensively source videos from multiple VQA databases \cite{qbenchvideo,qinstructvideo,lmmvqa}. 
However, due to the high cost of collecting MOS through annotated subjective studies, the scale of VQA databases is often only a fraction, ranging from one-tenth to even one-hundredth, of other visual task datasets \cite{adadqa,ptmvqa,qptv2}.
The limited number of videos in VQA databases hinders the content and quality diversity for further data scaling.
Thus, differentiating from prior studies, we identify \textbf{two key criteria} that allow us to diversify data collection from both labeled VQA and unlabeled general video databases, exemplified next.

\noindent \textbf{1) Number of annotated dimensions.}
For VQA databases, we argue that the quality richness of annotated scores in videos is more important than video quantity.
As summarized in Tab.\ref{tab: vqa}, compared to databases solely labeling a Mean Opinion Score (MOS) per video, each video in Maxwell \cite{maxwell} is labeled across 13 dimensions (\eg, noise, focus \etc).
All the scores are provided, reviewed, and averaged by 35 experts to ensure reliability.
Thus, we propose to use the number of annotated dimensions as the criterion for measuring labels' quality richness.
Adhering to this criterion, we only collect 4,543 \textit{labeled} videos from Maxwell, each with high quality richness in its annotations.

\noindent \textbf{2) Balanced quality distribution}.
Diversifying data collection by sampling videos from general video databases enables data scaling.
It also greatly enriches the semantics (\eg, scene and object categories) and temporal dynamics, thereby enhancing the quality diversity of the data \cite{ptmvqa,qpt}.
However, how to gather videos with balanced quality distribution without quality labels needs a solution.
To solve this, motivated by InternVid \cite{internvid}, we first leverage a lightweight video quality assessor \cite{AestheticPredictor} to efficiently calculate overall quality scores for video candidates from multiple databases as \textit{noisy quality labels}.
A uniform sampling procedure is then conducted to balance the quality distribution.
Since the noisy labels exhibit a discrepancy with the actual quality, we additionally filter out videos with incorrect labels, culminating in a selection of 100K \textit{unlabeled} videos. (\textit{More details in the Supp.})
As in Fig.\ref{fig:framework}.I, SIG collects a large scale of 104K videos with balanced quality distribution for upcoming automated scoring, depicted next.

\begin{figure*}[t]
  \centering
  \includegraphics[width=0.85\linewidth]{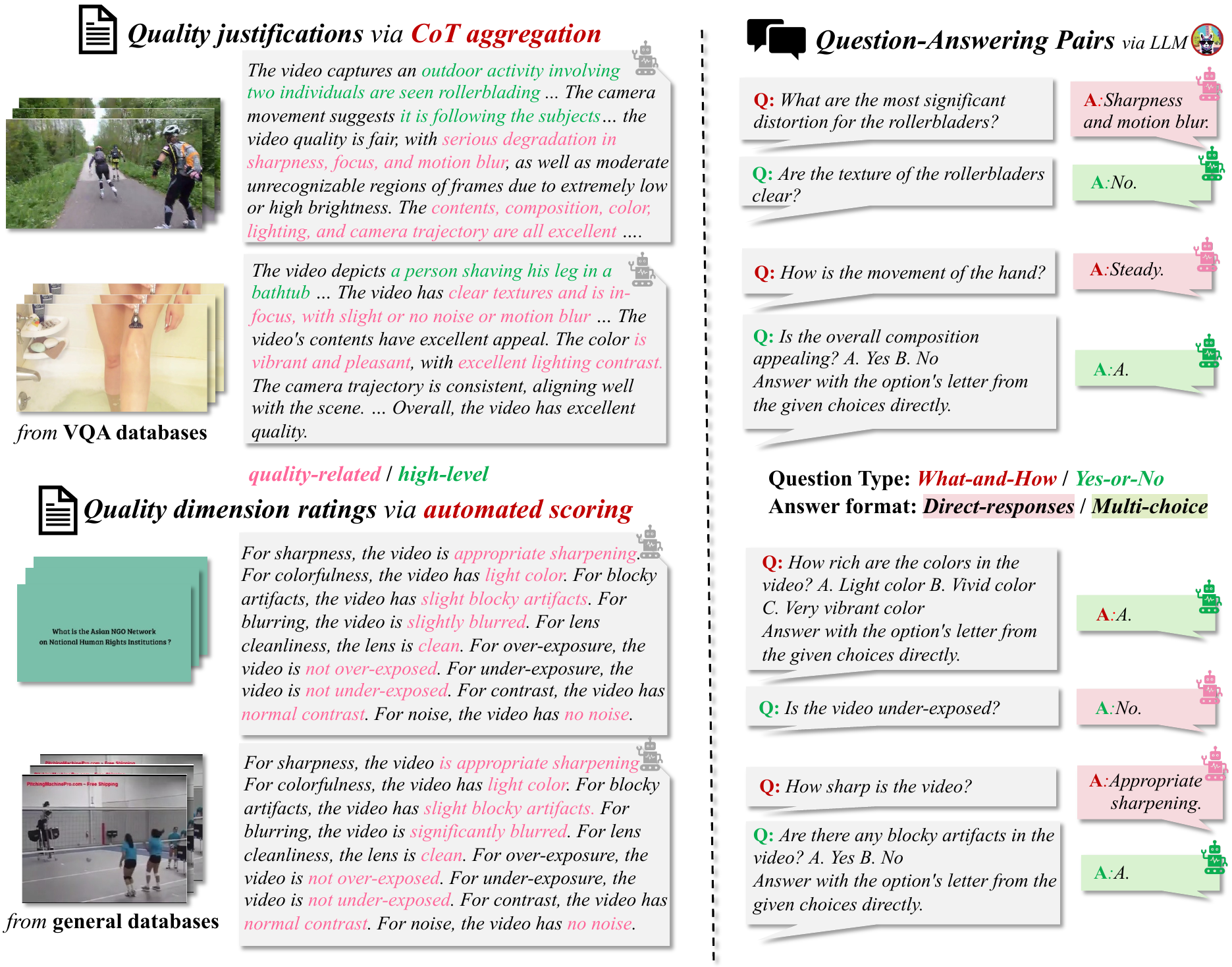}
  \caption{
  S2I comprises 320K instruction–response pairs: 104K detailed justifications with fine-grained ratings generated through automated scoring and CoT aggregation, and 216K question–answer pairs expanded by an LLM in both what-and-how and yes-or-no formats.
  }
  \label{fig:visualization}
\end{figure*}

\subsection{Automated Quality Dimension Scoring}
\label{method: sig_score}
After video source collection, a notable gap persists between unlabeled videos and quality-centric instructions. 
Since current LMMs generate imprecise instructions given their limited quality perception capacity \cite{qinstruct,mllmiqa}, and manually annotating massive videos is impractical.
Thus, we serve automated quality dimension scoring as the \textit{pivot} to connect videos and linguistic instructions, illustrated next.

\noindent \textbf{Selection of scoring dimensions.} 
To enhance the score's quality richness, we adhere to the proposed criterion by increasing the number of scoring dimensions.
To ensure the dimensions encompass all potential quality issues in the video, we rigorously enumerate them across the four stages of the video processing workflow (shooting, editing, compression, and transmission), ultimately selecting 14 dimensions (Fig.\ref{fig:framework}.II).
Our selection of quality dimensions is the most comprehensive to date compared to past studies \cite{mdvqa,kvq,finevq,maxwell}. \eg, we consider the \textit{clarity of the camera lens} during the \textit{shooting} phase. (\textit{More details in the Supp.})
Furthermore, to accurately score 14 quality dimensions for 100K videos, we resort to a practical video processing platform \cite{kvq} and invoke its deployed expert models, alleviating the subjective bias compared to crowdsourced scoring. All the quality scores are in the range of 0-1 for consistency.

\noindent \textbf{Mapping scores to text-defined levels.} 
The generated quality scores are inherently continuous, while LMMs generate discrete token outputs. 
Hence, we discretize scores to the standard five-tier text levels defined by ITU \cite{itu}, including \textit{bad}, \textit{poor}, \textit{fair}, \textit{good}, and \textit{excellent}.
We also find the quality dimension names exhibit ambiguity (\eg, the \textit{\underline{flicker}} is \textit{good}). 
To eliminate this ambiguity, we replace the dimension names with \textit{brief definitions}.
(\eg,  the \textit{\underline{variation smoothness between adjacent frames}} is \textit{good}. \textit{More details in the supp.} ) 
Moreover, the mapping operation inevitably results in quality information loss \cite{deqa}. 
Hence, we calculate the SRCC/PLCC between the text-defined levels and the original scores.
The computed SRCC and PLCC are above \textbf{0.95}, showing that the rating levels are sufficiently accurate.
Though a more accurate rating interval (\eg, seven-tier) might reduce the information loss, unlike the five-tier rating, its efficacy is not validated by human subjective study \cite{qalign}.
Overall, the automated quality dimension scoring makes it easy to scale up the dataset by annotating more unlabeled videos efficiently. 
Once each video's 14 quality dimensions are assigned text-defined ratings, SIG's final step is to aggregate them into complete quality justifications, described next.

\subsection{Hierarchical CoT Aggregation}
\label{method: sig_cot}
Complete quality justifications by humans encompass exhaustive descriptions of quality dimensions but, more importantly, incorporate the cognitive reasoning processes that derive the overall quality ratings given the ratings of individual quality dimensions \cite{hvs1,qinstruct}.
To mimic the HVS, we formulate this intuitive thought process into a hierarchical CoT, which considers the interplay of multiple interrelated quality issues for justification generation, exemplified next.

\noindent \textbf{CoT design.} 
As in Fig.\ref{fig:framework}.III, our CoT groups 14 quality dimensions into distortion- and aesthetic-related ones according to HVS's preference \cite{hvs2}.
It first evaluates the impact of each dimension and draws an intermediate rating in each group, then concludes a final quality evaluation given the two intermediate ratings.
This bottom-up, hierarchical scheme breaks down humans' quality perception into separate steps, enhancing the explainability of justifications.

\noindent \textbf{Refine justifications with high-level captions.} 
Albeit with reasoning, the resulting justifications exhibit identical format after CoT aggregation.
To enhance the naturalness of language, we first employ an open-source LLM (Vicuna-v1.5-7B \cite{vicuna}) for \textit{rephrasing} (Fig.\ref{fig:framework}.III).
Additionally, recent research shows that humans' quality perception often intertwines with high-level comprehension \cite{youtubeugc+,qptv2}. 
To supplement \textit{high-level content information} into quality justifications, a straightforward approach is to use off-the-shelf pretrained video captioners. 
Then, we employ ShareCaptioner-Video \cite{sharepgt4video} among the SOTA methods for its sliding-window scheme that preserves more detailed content information to obtain video captions. 
Last, the LLM gathers the high-level captions and rephrased justifications for final summarization. 
Since the randomness in LLM outputs may change the original text-defined rating levels \cite{gpt3}, producing erroneous responses. 
Thus, we mitigate this issue through prompt design, ensuring that the LLM does \textit{not} alter the rating levels during rephrasing and summarization processes, thereby safeguarding the accuracy of the quality information in final justifications. (\textit{Prompts for rephrasing and summarization are in the Supp.}) 
To this end, a total of \textbf{104K} video quality-centric justifications are developed, with rich quality issues and balanced quality distribution.

\noindent \textbf{Expand justifications to diverse instructions.}
Motivated by research in visual instruction tuning \cite{llava,llava1.5,vis_survey}, various instruction types help LMMs adeptly manage a broad array of real-world user queries for complex interactions.
Accordingly, we employ Vicuna-v1.5-7B \cite{vicuna} to generate QA pairs based on the generated quality justifications (Fig.\ref{fig:framework}.III). 
Following existing literature \cite{coinstruct,qinstruct}, our question forms include 'What/How' and 'Yes/No', while the answer formats are direct responses or multiple-choice questions (MCQ). 
We generate the distracting answer candidates on video quality, following the spirit of Q-Instruct \cite{qinstruct}.
After the LLM-assisted expansion, a total of \textbf{216K} QA pairs are acquired.
In brief, we conduct an in-depth analysis of the previous pipeline's bottleneck regarding data scalability, generation efficiency, and quality dimension coverage. 
By incorporating targeted improvements in the proposed SIG pipeline, we efficiently construct a large-scale dataset on video quality without human study or proprietary APIs. 
The dataset, termed S2Instruct, lays the foundation for later instruction tuning and benchmark curation, depicted next.

\section{The Score2Instruct and S2I-Bench}
\label{method: s2i_bench}

With S2I to boost quality scoring and justification capabilities, we first devise a progressive strategy tailored to data characteristics to exploit its potential. 
To better assess the quality justification capabilities of existing video LMMs, we then curate a benchmark with 400 challenging open-ended questions and four GPT-involved metrics.

\subsection{Score2Instruct}
\label{method: s2i}
\noindent \textbf{Dataset analysis.} 
As shown in Fig.\ref{fig:visualization}, 104K quality justifications in S2I encompass rich high-level descriptions and thorough quality issues that correlate well with the content. 
Besides, 216K question-answering pairs encompass diverse question types and response formats, underscoring the richness and accuracy of quality information.
For the instruction length statistics, 104K quality justifications have an average length of 129.4 words. 
For 216K question-answering pairs, there are 102K 'Yes/No' questions and 114K 'What/How' questions. 
The 'yes' and 'no' answers are balanced to a 1:1 ratio for a less biased evaluation \cite{qbench}. 

\noindent \textbf{Progressive tuning strategy.}
Recent advancements suggest various types of instructions enhance a model's distinct facets of instruction following abilities \cite{llavanextablation,vila}.
Following this spirit, we devise a two-stage tuning strategy to augment video LMMs' scoring and justification skills gradually.

\noindent \textit{\underline{Stage I Pretraining}}: 
Different from standard LMM pretraining, which leverages captions for vision-language feature alignment, stage I uses 100K dimension descriptions in S2I to predict text-defined quality ratings. 
Denote the video token as \texttt{<img>}, dimension name as \texttt{<dimension>}, dimension definition as \texttt{<definition>}, and rating level as \texttt{<level>}. 
The exemplar conversation format is as follows:

\noindent \textit{\#User:} \texttt{<img>} \textit{Rate the} \texttt{<dimension>} \textit{of the video.}

\noindent \textit{\#Assistant:} \textit{The} \texttt{<definition>} \textit{is} \texttt{<level>}.

\noindent Only the LMM responses (after \textit{\#Assistant:}) are supervised by the standard cross-entropy loss \cite{gpt3}.
Since the unified conversation format renders the task rather simple, we freeze the LLM and solely train the visual encoder and projector\footnote{Except for InternVL-Chat, we freeze its visual encoder due to the huge parameter count (6B) for efficiency.} to prevent overfitting.

\noindent \textit{\underline{Stage II Instruction Tuning}}:
Stage I endows models with initial quality scoring ability.
In stage II, we aim to achieve a more nuanced quality understanding and justification ability by training on 220K labeled justifications and QA pairs from S2I. 
At this stage, we adopt low-rank adaptation \cite{lora} on the LLM for efficiency (with a rank setting of $r=16$) and tune it with the connector and visual encoder. 
Stage II employs the cross-entropy loss \cite{gpt3} as in stage I. (\textit{Prompts and more details are in the Supp.})

\subsection{S2I-Bench}
\label{method: bench}

Existing benchmarks employ GPT to evaluate the reliability of the model's responses compared to the ground-truth quality justifications. It prompts GPT to conduct five repeated assessments to mitigate evaluation variance, outputting three metrics, including accuracy, completeness, and relevance. Each metric is rated on a scale from 0 to 2 \cite{qbench,qbenchvideo,llmjudge}.
We argue that \textbf{1)} these metrics fail to pay sufficient attention to video-specific aspects (\eg, temporal understanding), and \textbf{2)} writing ground-truth justifications from-scratch is highly nontrivial and costly.
(\textit{Details are in the Supp.})

To address \textbf{1)}, we propose to adapt VCG scores in \cite{videochatgpt} to video quality for a more well-rounded evaluation.
VCG scores include four video-specific metrics: Correctness of information (\textbf{CI}), detail orientation (\textbf{DO}), contextual understanding (\textbf{CU}), and temporal understanding (\textbf{TU}). 
Each score is on a scale of 1-5 and computed by prompting GPT-3 \cite{gpt3}. (\textit{Prompts are in the Supp.}) 
These improved metrics allow us to discover more insights from the experimental results.
To address \textbf{2)}, we uniformly sample 400 video-justification pairs from S2I (nonoverlap with the training set) to curate the challenging benchmark S2I-Bench.
To ensure the benchmark's reliability, we conduct a thorough manual check, correcting all the errors in the automatically annotated quality justifications.

\begin{table*}[t]
\belowrulesep=0pt
\aboverulesep=0pt
\centering
\caption{Performance benchmarking of baseline models and S2I-tuned models on proposed S2I-Bench. Metrics are VCG scores.}
\vspace{-2pt}
\label{tab:description}
\begin{adjustbox}{max width=0.8\textwidth}
\begin{tabular}{cc|cccc|c}
\toprule
\multicolumn{1}{c:}{\textbf{Model} (\textit{LLM Variants})} & \textit{S2I-Tuned} & \textit{CI}$\uparrow$ & \textit{CU}$\uparrow$ & \textit{DO}$\uparrow$ & \textit{TU}$\uparrow$ & \textbf{\textit{Sum}}$\uparrow$ \\ 
\hline
\multicolumn{1}{c:}{\multirow{2}{*}{LLaVA-OneVision (\textit{Vicuna-v1.1-7B})}} & \textit{No} (Baseline) & 2.11 & 2.30 & 1.83 & 1.72 & 7.96 \\
\multicolumn{1}{c:}{} & \textbf{\textit{Yes}} & 3.02\textsubscript{\textcolor{green}{+0.91}} & 2.49\textsubscript{\textcolor{green}{+0.19}} & 2.13\textsubscript{\textcolor{green}{+0.30}} & 2.24\textsubscript{\textcolor{green}{+0.72}} & 9.88\textsubscript{\textcolor{green}{+1.92}} \\ \hdashline
\multicolumn{1}{c:}{\multirow{2}{*}{LLaVA-Next-Video (\textit{Mistral-7B})}} & \textit{No} (Baseline) & 2.08 & 2.19 & 1.77 & 1.64 & 7.68 \\
\multicolumn{1}{c:}{} & \textbf{\textit{Yes}} & 2.43\textsubscript{\textcolor{green}{+0.35}} & 2.35\textsubscript{\textcolor{green}{+0.16}} & 2.16\textsubscript{\textcolor{green}{+0.39}} & 2.07\textsubscript{\textcolor{green}{+0.43}} & 
9.01\textsubscript{\textcolor{green}{+1.33}}\\ 
\hdashline
\multicolumn{1}{c:}{\multirow{2}{*}{InternVL-Chat (\textit{Vicuna-7B})}} & \textit{No} (Baseline) & 2.19 & 2.26 & 1.90 & 1.67 & 8.02 \\
\multicolumn{1}{c:}{} & \textbf{\textit{Yes}} & 2.68\textsubscript{\textcolor{green}{+0.49}} & 2.74\textsubscript{\textcolor{green}{+0.48}} & 2.11\textsubscript{\textcolor{green}{+0.21}} & 2.19\textsubscript{\textcolor{green}{+0.52}} &  
9.72\textsubscript{\textcolor{green}{+1.70}}
\\ \hdashline
\multicolumn{1}{c:}{\multirow{2}{*}{Video-LLaVA (\textit{Vicuna-v1.5-7B})}} & \textit{No} (Baseline) & 2.03 & 2.06 & 1.60 & 1.46 & 7.15 \\
\multicolumn{1}{c:}{} & \textbf{\textit{Yes}} & 2.28\textsubscript{\textcolor{green}{+0.25}} & 2.37\textsubscript{\textcolor{green}{+0.31}} & 1.97\textsubscript{\textcolor{green}{+0.37}} & 1.88\textsubscript{\textcolor{green}{+0.42}} &  
8.50\textsubscript{\textcolor{green}{+1.35}} \\ \hdashline
\multicolumn{1}{c:}{\multirow{2}{*}{LLaVA-Video (\textit{Qwen2-7B})}} & \textit{No} (Baseline) & 2.05 & 2.08 & 2.06 & 1.98 & 8.17 \\
\multicolumn{1}{c:}{} & \textbf{\textit{Yes}} & 2.22\textsubscript{\textcolor{green}{+0.17}} & 2.30\textsubscript{\textcolor{green}{+0.22}} & 2.25\textsubscript{\textcolor{green}{+0.19}} & 2.31\textsubscript{\textcolor{green}{+0.33}} &  
9.08\textsubscript{\textcolor{green}{+0.91}}\\ \hdashline
\multicolumn{1}{c:}{\multirow{2}{*}{VideoLLaMA3 (\textit{Qwen2.5-7B})}} & \textit{No} (Baseline) & 2.14 & 2.14 & 1.97 & 2.06 & 8.31 \\
\multicolumn{1}{c:}{} & \textbf{\textit{Yes}} & 2.38\textsubscript{\textcolor{green}{+0.24}} & 2.28\textsubscript{\textcolor{green}{+0.14}} & 2.25\textsubscript{\textcolor{green}{+0.28}} & 2.36\textsubscript{\textcolor{green}{+0.30}} &  
9.27\textsubscript{\textcolor{green}{+0.96}}
\\ \hline
\multicolumn{2}{c|}{\textit{Average Improvements}} & +\textit{0.40} & +\textit{0.25} & +\textit{0.29} & +\textit{0.45} & +\textit{1.39} \\ 
\bottomrule
\end{tabular}
\end{adjustbox}
\end{table*}
\begin{table*}[t]
\belowrulesep=0pt
\aboverulesep=0pt
\centering
\caption{SRCC\&PLCC of baseline and S2I-tuned models on five in-the-wild VQA datasets.
Best results in \textbf{bold}.}
\vspace{-2pt}
\label{tab:acc}
\begin{adjustbox}{max width=0.95\textwidth}
\begin{tabular}{cc|c:cccc}
\toprule
\multicolumn{2}{c|}{\textbf{Dataset Group}} & \textit{Intra-dataset} & \multicolumn{4}{c}{\textit{Cross-dataset}} \\ \hdashline
\multicolumn{1}{c:}{\textbf{Model} (\textit{LLM Variants})} & \textit{S2I-Tuned} & \textit{Maxwell} &\textit{ LSVQ$_{test}$} & \textit{LSVQ$_{1080p}$} & \textit{KoNViD-1k} & \textit{LIVE-VQC}
\\ \hline
\multicolumn{1}{c:}{\multirow{2}{*}{LLaVA-OneVision (\textit{Vicuna-v1.1-7B})}} & \textit{No} (Baseline) & 0.474 / 0.428 & 0.449 / 0.438 & 0.337 / 0.311 & 0.392 / 0.394 & 0.397 / 0.410 \\
\multicolumn{1}{c:}{} & \textbf{\textit{Yes}} & 0.795 / 0.812 & 0.751 / 0.730 & 0.671 / 0.634 & 0.726 / 0.689 & 0.738 / 0.752
\\ \hdashline
\multicolumn{1}{c:}{\multirow{2}{*}{LLaVA-Next-Video (\textit{Mistral-7B})}} & \textit{No} (Baseline) & 0.503 / 0.457 & 0.509 / 0.517 & 0.362 / 0.353 & 0.477 / 0.440 & 0.406 / 0.433  \\
\multicolumn{1}{c:}{} & \textbf{\textit{Yes}} &  \textbf{0.847} / 0.795 & 0.743 / 0.727 & 0.675 / 0.705 & 0.624 / 0.636 & 0.526 / 0.558  
\\ \hdashline 
\multicolumn{1}{c:}{\multirow{2}{*}{InternVL-Chat (\textit{Vicuna-7B})}} & \textit{No} (Baseline) & 0.358 / 0.302 & 0.394 / 0.331 & 0.289 / 0.316 & 0.365 / 0.323 & 0.347 / 0.366  \\
\multicolumn{1}{c:}{} & \textbf{\textit{Yes}} &  0.703 / 0.723 & 0.621 / 0.636 & 0.582 / 0.590 & 0.597 / 0.642 & 0.574 / 0.508
\\ \hdashline
\multicolumn{1}{c:}{\multirow{2}{*}{Video-LLaVA (\textit{Vicuna-v1.5-7B})}} & \textit{No} (Baseline) & 0.451 / 0.394 & 0.412 / 0.383 & 0.294 / 0.308 & 0.430 / 0.434 & 0.461 / 0.445  \\
\multicolumn{1}{c:}{} & \textbf{\textit{Yes}} &  0.808 / \textbf{0.834} & 0.687 / 0.692 & 0.680 / 0.653 & 0.642 / 0.694 & 0.673 / 0.671  \\ \hdashline
\multicolumn{1}{c:}{\multirow{2}{*}{LLaVA-Video (\textit{Qwen2-7B}))}} & \textit{No} (Baseline) & 0.564 / 0.557 & 0.494 / 0.446 & 0.422 / 0.380 & 0.535 / 0.488 & 0.572 / 0.530  \\
\multicolumn{1}{c:}{} & \textbf{\textit{Yes}}  & 0.826 / 0.774 & 0.760 / 0.734 & 0.667 / 0.652 & \textbf{0.773} / \textbf{0.769} & 0.730 / \textbf{0.765}
\\ \hdashline
\multicolumn{1}{c:}{\multirow{2}{*}{VideoLLaMA3 (\textit{Qwen2.5-7B})}} & \textit{No} (Baseline) & 0.548 / 0.516 & 0.533 / 0.486 & 0.483 / 0.454 & 0.592 / 0.539 & 0.579 / 0.547  \\
\multicolumn{1}{c:}{} & \textbf{\textit{Yes}} & 0.801 / 0.761 & \textbf{0.793} / \textbf{0.788} & \textbf{0.705} / \textbf{0.714} & 0.749 / 0.695 & \textbf{0.763} / 0.742
\\
\hline
\multicolumn{2}{c|}{\textit{Average Improvements}} & \textcolor{green}{+\textit{0.314}}\textit{/}\textcolor{green}{+\textit{0.340}} & \textcolor{green}{+\textit{0.261}}\textit{/}\textcolor{green}{+\textit{0.284}} & \textcolor{green}{+\textit{0.299}}\textit{/}\textcolor{green}{+\textit{0.304}} & \textcolor{green}{+\textit{0.220}}\textit{/}\textcolor{green}{+\textit{0.251}} & \textcolor{green}{+\textit{0.207}}\textit{/}\textcolor{green}{+\textit{0.211}}  \\
\bottomrule
\end{tabular}
\end{adjustbox}
\end{table*}

\section{Experiments}
\label{exp}

\subsection{Experimental Settings}
\label{exp: settings}

\noindent \textbf{Baseline models.}
We select six video LMMs to evaluate their video quality scoring and justification abilities \textit{before} and \textit{after} tuning on S2I.
The models include LLaVA-OneVision \cite{llavaonevision}, LLaVA-Next-Video \cite{llavanextvideo}, InternVL-Chat \cite{internvlchat}, Video-LLaVA \cite{videollava}, LLaVA-Video \cite{llavavideo}, VideoLLaMA3 \cite{videollama3}. 
All models are trained on the respective subset of S2I for one epoch in both tuning stages, and we only use the 7B variants for a fair comparison. 
16 frames are uniformly sampled for evaluation.
Zero-shot performances of three closed-source LMMs (GPT-4o \cite{gpt4o}, GPT-4o-mini \cite{gpt4omini}, and Gemini-1.5 Pro \cite{gemini1.5pro}) and more results are in the \textit{Supp}.

\noindent \textbf{Benchmarks and criteria.}
We evaluate the quality justification ability of video LMM on the introduced S2I-Bench. 
All responses from video LMMs are generated with greedy search. 
To greenuce the inference variance on the S2I-Bench, we evaluate the same LMM five times and average the metrics for the final results.
We calculate SRCC (Spearman rank correlation coef.) and PLCC (Pearson linear correlation coef.) on five standard in-the-wild VQA datasets, including Maxwell \cite{maxwell}, LSVQ$_{test}$ \cite{pvq}, LSVQ$_{1080p}$ \cite{pvq}, KoNViD-1k \cite{kv1k}, and LIVE-VQC \cite{livevqc} to evaluate the visual scoring ability. 
A larger SRCC indicates a better ranking between samples, and a larger PLCC shows a more accurate score pgreeniction.
We follow the softmax pooling operation in \cite{qalign} to generate the final quality score.
The evaluation prompts of quality scoring and justification are in the \textit{supp}.

\subsection{Main Results}
\label{exp: sota}

\noindent \textbf{Video quality justification.} 
We compare the results on S2I-Bench before and after the tuning on S2I.
As in Tab.\ref{tab:description}, VCG scores of the six baseline models exhibit significant enhancements, with the average summation of scores increasing by \textit{1.39}. 
This finding substantiates the efficacy of training on S2Istruct. 
Among these, the CI and TU metrics exhibit the most notable improvements, reflecting that the trained model provides a more comprehensive portrayal of quality issues and demonstrates heightened sensitivity to temporal-related concerns. 
Moreover, the improvement of CU and DO demonstrates that model performance post-training is more adept at capturing nuances. 
We expect further to scale the S2I-Bench in the future for more insights.

\noindent \textbf{Video quality scoring.} 
In Tab.\ref{tab:acc}, the SRCC and PLCC of baseline models both show improvement, thereby demonstrating the efficacy of S2I in enhancing scoring accuracy, even in the absence of direct supervision through numerical scores.
The advancement of the intra-dataset scenario is most pronounced across five datasets, indicating that the SIG pipeline effectively harnessed the potential of the quality dimension ratings.
More remarkably, there also exist large gains in cross-dataset scenario (\eg, \textbf{+29.9\%} of SRCC and \textbf{+30.4\%} of PLCC for LSVQ$_{1080p}$). 
This indicates that despite a notable distribution shift between VQA datasets, our SIG pipeline can exploit diverse quality factors for training, thereby enhancing models' generalizability.

\subsection{Ablation and Analyses}
\label{exp: ablation}
\begin{table}[t]
\belowrulesep=0pt
\aboverulesep=0pt
\centering
\caption{Ablation on the video source collection. Metrics are VCG scores and SRCC/PLCC. 
}
\label{tab:ablation_video}
\scalebox{0.73}{
\begin{tabular}{c|cccc:cc}
\toprule
\textbf{Video}  & \textit{CI}$\uparrow$ & \textit{CU}$\uparrow$ & \textit{DO}$\uparrow$ & \textit{TU}$\uparrow$ & \textit{Maxwell} & \textit{LIVE-VQC} 
\\ \hline
\rowcolor{grey}\textit{\textbf{w/ All}} (adopted) & 3.02 & 2.49 & 2.13 & 2.24 & 0.795/0.812 & 0.738/0.752 \\
\textit{w/o labeled} & 2.44 & 1.76 & 1.68 & 1.59 & 0.738/0.702 & 0.663/0.607\\
\textit{w/o unlabeled} & 2.57 & 2.08 & 2.10 & 2.14 & 0.506/0.553 & 0.467/0.418 \\
\bottomrule
\end{tabular}}
\end{table}
\begin{table}[t]
\belowrulesep=0pt
\aboverulesep=0pt
\centering
\caption{Ablation on the key designs in CoT. 
\textbf{a)} the hierarchical design, \textbf{b)} the high-level caption refinement, and \textbf{c)} the prompt design to keep rating levels unchanged.
}
\label{tab:ablation_aggregation}
\scalebox{0.73}{
\begin{tabular}{c|cccc:cc}
\toprule
\textbf{CoT}  & \textit{CI}$\uparrow$ & \textit{CU}$\uparrow$ & \textit{DO}$\uparrow$ & \textit{TU}$\uparrow$ & \textit{Maxwell} & \textit{LIVE-VQC}
\\ \hline
\rowcolor{grey}\textbf{\textit{w/ All}} (adopted) & 3.02 & 2.49 & 2.13 & 2.24 & 0.795/0.812 & 0.738/0.752 
\\
\textit{w/o \textbf{a)}} & 2.96 & 2.41 & 2.08 & 2.16 & 0.786/0.810 & 0.724/0.737 \\ 
\textit{w/o \textbf{b)}} & 2.54 & 2.35 & 1.98 & 2.09 & 0.750/0.723 & 0.712/0.706 \\ 
\textit{w/o \textbf{c)}} & 3.02 & 2.49 & 2.13 & 2.24 & 0.604/0.658 & 0.573/0.540 \\ 
\bottomrule
\end{tabular}}
\end{table}
\begin{table}[t]
\belowrulesep=0pt
\aboverulesep=0pt
\centering
\caption{Ablation on the progressive tuning strategy. Metrics are VCG scores and SRCC/PLCC. 
}
\label{tab:ablation_pretrain}
\scalebox{0.77}{
\begin{tabular}{c|cccc:cc}
\toprule
\textbf{Pretraining}  & \textit{CI}$\uparrow$ & \textit{CU}$\uparrow$ & \textit{DO}$\uparrow$ & \textit{TU}$\uparrow$ & \textit{Maxwell} & \textit{LIVE-VQC} 
\\ \hline
\rowcolor{grey}\textbf{\textit{w/}} (adopted) & 3.02 & 2.49 & 2.13 & 2.24 & 0.795/0.812 & 0.647/0.683 \\
\textit{w/o} & 2.86 & 2.49 & 2.10 & 2.19 & 0.638/0.592 & 0.738/0.752\\
\bottomrule
\end{tabular}}
\end{table}
\begin{table}[t]
\belowrulesep=0pt
\aboverulesep=0pt
\caption{Ablation on the data scalability of SIG. Metrics are VCG scores and SRCC/PLCC.
}
\centering
\label{tab:ablation_scaling}
\scalebox{0.73}{
\begin{tabular}{c|cccc:cc}
\toprule
\textbf{Precentage}  & \textit{CI}$\uparrow$ & \textit{CU}$\uparrow$ & \textit{DO}$\uparrow$ & \textit{TU}$\uparrow$ & \textit{Maxwell} & \textit{LIVE-VQC}
\\ \hline
\textit{20\%} & 2.27 & 2.31 & 1.86 & 1.80 & 0.623/0.637 & 0.475/0.492 \\
\textit{50\%} & 2.44 & 2.38 & 2.04 & 1.95 & 0.694/0.715 & 0.556/0.537 \\
\rowcolor{grey}\textbf{\textit{100\%}} (adopted) & 3.02 & 2.49 & 2.13 & 2.24 & 0.795/0.812 & 0.738/0.752  \\ 
\bottomrule
\end{tabular}}
\end{table}
\vspace{-6pt}

In this section, all the ablation experiments are conducted on LLaVA-OneVision \cite{llavaonevision} if not mentioned.

\noindent \textbf{Effects of the video source collection.}
As in Tab.\ref{tab:ablation_video}, the metrics training solely on labeled or unlabeled videos are inferior to those on the entire S2I.
The limited labeled data contributes more to quality justifications, while the massive unlabeled data boosts quality scoring more, underscoring the effectiveness of the proposed data sourcing criteria.

\noindent \textbf{Effectiveness of hierarchical CoT aggregation.} 
Several findings can be drawn from Tab.\ref{tab:ablation_aggregation}: 
\textbf{1)} The hierarchical design mainly influences the justification ability (the total decline across the four metrics is 0.27). 
Given the scarcity of reliable quality dimension annotations, such as those in Maxwell, we consider employing automated scoring 
on a broader array of videos in future research.
\textbf{2)} The high-level captions impact more on the quality justification, further highlighting the correlation between quality and content. 
The notable declines in CU, DO, and TU indicate that LMMs struggle to associate quality issues with the corresponding content without high-level awareness. 
Instead, using the off-the-shelf video captioner can alleviate this problem efficiently.
\textbf{3)} The design c) affects scoring accuracy more, indicating that the discretized levels of scores serve as the foundation for the ability to rate quality issues. 

\noindent \textbf{Effectiveness of progressive tuning strategy.} 
Beyond the conventional instruction tuning phase on various instructions, the hallmark of our strategy is incorporating a preceding pretraining stage.
As shown in Tab.\ref{tab:ablation_pretrain}, the removal of stage I results in a degradation of the justification and scoring abilities. 
Particularly, pretraining exerts a more significant influence on scoring (\eg, \textbf{-15.7\%} of SRCC and \textbf{-22.0\%} of PLCC for Maxwell), suggesting that although the concise, template-like dimension-wise descriptions lack language diversity, they enable LMMs to concentrate on the evaluation across quality dimensions.
This finding prompts us to consider whether incorporating more quality dimensions other than visual effects in pretraining data could further enhance scoring accuracy (\eg, the audio quality of the video). 
We leave this to the future work.

\noindent \textbf{Scalability of automated scoring.}
Our primary insight into SIG lies in the accessibility and scalability of quality scores. 
To confirm the scalability of the SIG pipeline, we use 20\%, 50\%, and 100\% percentages of the training data \textit{generated by automated scoring} (\eg, quality dimension ratings). 
As in Tab.\ref{tab:ablation_scaling}, scaling up data via automated scoring during training can continuously improve the justification and scoring abilities. 
Also, the performances are not saturated, given the current data scale.
Thus, automatically scoring more videos via our SIG pipeline might potentially gain better results and could be explored in future work.

\section{Conclusion}
\label{conclusion}

This paper bridges video quality scoring and justification of video LMMs via quality-centric instruction tuning. 
The score-based instruction generation (SIG) pipeline is crafted to efficiently generate scalable data, overcoming the constraints imposed by VQA databases, human annotation costs, and proprietary system usage.
By progressively tuning on the constructed Score2Instruct (S2I) dataset with over 320K diverse instruction-response pairs, we show that multiple video LMMs demonstrate advanced quality scoring and justification abilities simultaneously, surpassing the baselines and closed-source models.
Further, we derive the S2I-Bench for more comprehensive analyses, making a solid step in benchmarking quality understanding and justification abilities for video LMMs.
In summary, we hope our data-centric perspective can inspire the community to broaden the scope of VQA and contribute to developing versatile quality assessors in future research.
\clearpage
{
    \small
    \bibliographystyle{ieeenat_fullname}
    \bibliography{main}
}
\clearpage

\setcounter{equation}{0}
\setcounter{figure}{0}
\setcounter{table}{0}
\setcounter{section}{0}
\renewcommand{\theequation}{\arabic{equation}}
\renewcommand{\thefigure}{\arabic{figure}}
\renewcommand{\thetable}{\arabic{table}}

\twocolumn[
\begin{center}
    {\Large \textbf{Score2Instruct: Scaling Up Video Quality-Centric Instructions \\via Automated Dimension Scoring \\ \vspace{0.5em} \textit{Supplementary Material}} \par}
    \vspace{1.5em}
    {\large
      \begin{tabular}[t]{c}
        Qizhi Xie$^{1,2}$, Kun Yuan$^{2\textrm{ \Letter}}$, Yunpeng Qu$^{1,2}$, Jiachao Gong$^2$, Mingda Wu$^2$\\, Ming Sun$^2$, Chao Zhou$^2$, Jihong Zhu$^{1\textrm{ \Letter}}$ \\
        $^1$Tsinghua University, $^2$Kuaishou Technology \\
        \texttt{\small xqz20@mail.tsinghua.edu.cn, yuankun03@kuaishou.com, jhzhu@tsinghua.edu.cn}
      \end{tabular}\par
    }
    \vspace{2em}
\end{center}
]

\section{More Experimental Results}
\label{supp:compare}
\subsection{More Results on Q-Bench-Video}
We further evaluate on \textit{Q-Bench-Video} (Tab.\ref{tab:qbenchvideo}), on which the S2I-tuned model also shows a notable performance gain.

\begin{table}[t]
\centering
\caption{Q-Bench-Video evaluation (before/after) S2I tuning. Metrics are SRCC and PLCC.}
\label{tab:qbenchvideo}
\scalebox{0.7}{
\begin{tabular}{c|cccc}
\toprule
\textbf{Model} & \textit{Tech.}$\uparrow$ & \textit{Aes.}$\uparrow$ & \textit{Temp.}$\uparrow$ & \textit{AIGC}$\uparrow$ \\ 
\hline
LLaVA-OV-7B & (0.493/0.562) & (0.641/0.553) & (0.506/0.528) & (0.443/0.487) \\
InternVL-7B & (0.484/0.531) & (0.527/0.550) & (0.505/0.535) & (0.531/0.537)\\
\bottomrule
\end{tabular}}
\end{table}
\subsection{Comparison with More Methods}
The core contribution of this paper is proving the efficacy of automated scoring to scale up quality instructions.
We add more comparisons on quality justification and scoring tasks (Tab.\ref{tab:just} and Tab.\ref{tab:score}). 
The S2I-tuned models excel in the justification task. 
Although the scoring performance is inferior, our paper focuses on a more challenging setting: Leveraging massive in-the-wild videos to break the annotation barrier, and output justifications spanning comprehensive dimensions. 
We'll scale up the data and incorporate tailored model designs to boost the performance in future work.

\begin{table}[t]
    \centering
    \caption{Comparison on quality justification task with more methods, using LLaVA-OneVision.}
    \scalebox{0.8}{
    \begin{tabular}{c|cccc}
    \toprule
        \textbf{Model} & \textit{CI}$\uparrow$ & \textit{CU}$\uparrow$ & \textit{DO}$\uparrow$ & \textit{TU}$\uparrow$\\ 
        \hline
        LLaVA-OV-7B (\textit{Ours}) & 3.02 & 2.49 & 2.13 & 2.24 \\
        Q-Instruct & 1.82 & 1.47 & 1.95 & 1.83 \\
        Depict-QA & 1.74 & 1.55 & 2.04 & 1.24 \\
        Qwen2.5-VL-32B & 3.12 & 2.28 & 2.10 & 2.15 \\
        \bottomrule
    \end{tabular}}
    \label{tab:just}
\end{table}
\begin{table}[t]
\centering
\caption{Comparison on quality scoring task with more methods, using VideoLLaMA3.}
\label{tab:score}
\scalebox{0.58}{
\begin{tabular}{c|ccccc}
\toprule
\textbf{Model} & \textit{Maxwell} & \textit{LSVQ$_{test}$} & \textit{LSVQ$_{1080p}$} & \textit{KoNViD-1k} & \textit{LIVE-VQC}\\ 
\hline
VideoLLaMA3 (\textit{Ours}) & 0.801/0.761 & 0.793/0.788 & 0.705/0.714 & 0.749/0.695 & 0.763/0.742 \\
Q-Align & 0.780/0.782 & 0.883/0.882 & 0.797/0.830 & 0.865/0.877 & 0.847/0.832 \\
Fast-VQA & 0.720/0.728 & 0.876/0.877 & 0.779/0.814 & 0.859/0.855 & 0.823/0.844\\
PVQ & 0.698/0.703 & 0.814/0.816 & 0.686/0.708 & 0.781/0.781 & 0.747/0.776\\
\bottomrule
\end{tabular}}
\end{table}
\subsection{Compare S2I with Human-Annotated Instruction Dataset}
We keep the same progressive training strategy in \textit{Score2Instruct} section by tuning on the \textit{Stage-2\&3 dataset} proposed by VQA$^2$~\cite{qinstructvideo}. As in Tab.\ref{tab:human}, the results prove the advantage of machine-annotated S2I beyond its cost efficiency.

\begin{table}[t]
\centering
\caption{Comparison with human-annotated dataset, using LLaVA-OneVision.}
\label{tab:human}
\scalebox{0.73}{
\begin{tabular}{c|cccc|cc}
\toprule
\textbf{Dataset}  & \textit{CI}$\uparrow$ & \textit{CU}$\uparrow$ & \textit{DO}$\uparrow$ & \textit{TU}$\uparrow$ & \textit{Maxwell} & \textit{LIVE-VQC}
\\ \hline
S2I (\textit{Ours})  & 3.02 & 2.49 & 2.13 & 2.24 & 0.795/0.812 & 0.738/0.752 
\\ 
VQA$^2$ & 2.45 & 2.30 & 1.96 & 2.05 & 0.746/0.723 & 0.720/0.716 \\
\bottomrule
\end{tabular}}
\end{table}
\subsection{Ablation on Model Architecture}
We uniform sample 16 frames for evaluation, as the model architecture design is not our main focus.
Yet, we add results using the \textit{slow-fast} sampling strategy in FineVQ~\cite{finevq} for tuning (Tab.\ref{tab:sampling}), better capturing temporal quality issues.

\begin{table}[t]
\centering
\caption{Comparison with sampling strategy, using LLaVA-OneVision.}
\label{tab:sampling}
\scalebox{0.73}{
\begin{tabular}{c|cccc|cc}
\toprule
\textbf{Sampling}  & \textit{CI}$\uparrow$ & \textit{CU}$\uparrow$ & \textit{DO}$\uparrow$ & \textit{TU}$\uparrow$ & \textit{Maxwell} & \textit{LIVE-VQC}
\\ \hline
Uniform (Ours)  & 3.02 & 2.49 & 2.13 & 2.24 & 0.795/0.812 & 0.738/0.752 
\\ 
Slow-fast & 3.05 & 2.66 & 2.25 & 2.37 & 0.802/0.813 & 0.795/0.808 \\
\bottomrule
\end{tabular}}
\end{table}

\section{Novelty Clarification}
We clarify our novelty below. Compared to FineVQ~\cite{finevq}, SIG 1) eliminates the need for \textit{expert} scoring and \textit{proprietary} APIs, 2) offers greater scalability via exploiting \textit{unlabelled} videos, and 3) covers \textit{more} dimensions. 

Compared to VQA$^2$~\cite{qinstructvideo}, our main contributions lie not in the training strategy but in SIG, S2I, and S2I-Bench, and we do \textit{not} require separate models for the two tasks.

\section{Details of the Score2Instruct}
\label{supp:s2i}

\subsection{Descriptions of Quality Dimensions}
\label{supp:s2i_score}

In \textit{Automated Quality Dimension Scoring} section, we enumerate a total of 14 quality dimensions to cover all the quality issues that might appear in the video.
All the dimensions are scored on a scale of 0-1.
Here, we provide a detailed definition of each dimension.

\begin{itemize}
    \item \textbf{Focus}: The probability of the salient target in the video is in focus and not looking Gaussian-blurred.
    \item \textbf{Clarity of camera lens}: The probability of no blemishes or smudges on the camera lens. 
    \item \textbf{Exposure}: The probability of no unrecognizable regions of frames due to extremely low or high brightness.
    \item \textbf{Noise}: The probability of no random pixel-wise brightness or color variation.
    \item \textbf{Sharpness}: The probability of not having clear textures.
    \item \textbf{Compression artifacts}: The probability of not having block-like or moire-like artifacts introduced by compression algorithms.
    \item \textbf{Motion blur}: The probability of not having blurriness that happens during and is caused by the motions of camera or subjects in the video.
    \item \textbf{Fluency}: The probability of no missing frames during a moving sequence.
    \item \textbf{Flicker}: The probability of no non-smooth variation between adjacent frames.
    \item \textbf{Camera trajectory}: The probability of the camera moving in a consistent temporal trajectory that aligns with the scene.
    \item \textbf{Contrast}: The probability of having proper contrastive lighting in the video.
    \item \textbf{Content complexity}: The probability of having a rich diversity of textures.
    \item \textbf{Content composition}: The probability of having an organized and balanced composition of objects and scenes.
    \item \textbf{Colorfulness}: The probability of having vibrant and pleasant color.
\end{itemize}

\noindent Each dimension is scored by an expert model in \cite{kvq} platform.
Each expert model is trained on large-scale UGC videos and verified in \cite{kvq}.
After scoring and mapping to discrete text-defined levels, the quality dimension rating is obtained by concatenating dimension definition and level. 

\subsection{Ablation on Quality Dimensions}
We ablate \textit{distortion} and \textit{aesthetic} dimensions (Tab.\ref{tab:dimension}). The distortion dimensions hold slightly greater importance.

\begin{table}[t]
\centering
\caption{Ablation on quality dimensions, including distortion and aesthetic dimensions.}
\label{tab:dimension}
\scalebox{0.73}{
\begin{tabular}{c|cccc|cc}
\toprule
\textbf{Dimension}  & \textit{CI}$\uparrow$ & \textit{CU}$\uparrow$ & \textit{DO}$\uparrow$ & \textit{TU}$\uparrow$ & \textit{Maxwell} & \textit{LIVE-VQC}
\\ \hline
Distortion only  & 2.96 & 2.49 & 2.12 & 2.17 & 0.743/0.752 & 0.706/0.688 
\\ 
Aesthetic only & 2.80 & 2.44 & 2.07 & 2.19 & 0.733/0.727 & 0.686/0.674 \\
\bottomrule
\end{tabular}}
\end{table}

\subsection{Details of Expert Models}
\label{supp:s2i_kvq}
The architecture is based on ConvNeXt, and a model is trained on an expert-labelled MOS dataset for each dimension for high accuracy.

\subsection{Human Filtering in Video Source Collection}
\label{supp:s2i_video}

In \textit{Video Source Collection} section, we leverage a lightweight video quality assessor \cite{AestheticPredictor} to gain noisy quality labels. 
To filter out erroneous labels, we conduct a filtering process as follows.
The \cite{AestheticPredictor} is built on CLIP \cite{clip} by fine-tuning a linear layer (linear probing) on IQA data.
Due to CLIP's tendency to assign extreme aesthetic scores \cite{clipiqa}, we review 2K videos with the highest and lowest aesthetic scores, excluding those with inaccurate scores.

In all, the synthetic data aligns well with humans because 1) the \textit{scoring models} and proposed \textit{CoT} align well with experts, 2) the \textit{discrete ratings} follow the ITU standard. We also force the LLM \textit{not} to change the ratings by prompt design (See \textit{Prompt Design in Progressive Tuning} section) to avoid bias propagation.

\subsection{Details of QA generation}
\label{supp:s2i_qa}
The diversity and correctness are secured by \textit{curated question and answer sets} for each dimension. 
We prompt the LLM to generate 50 questions, from which we eliminate repetitive and erroneous ones, resulting in 20 questions.
The answer set is transformed and rephrased from the five-tier text ratings to minimise hallucination.

\subsection{Subjectivity Discussion}
\label{supp:s2i_sub}

We note that subjectivity in human annotation is unavoidable. 
Conversely, the scoring models offer \textit{better consistency} compared to humans, and we only use justifications unanimously approved by \textit{all raters} in S2I and S2I-Bench to minimise potential subjectivity.
\subsection{Prompt Design}
\label{supp:s2i_prompt}

In \textit{Hierarchical CoT Aggregation} section, an open-source LLM Vicuna-v1.5-7B \cite{vicuna} is employed to rephrase and summarize the quality justifications. 
The prompts are as follows.

\subsubsection{Rephrasing}
\label{supp:s2i_prompt_s}

\textit{\#User: 
I will provide you with a text on video quality assessment that reflects the reasoning process for evaluating video quality. 
I need you to rephrase this text. Please note: 
1. The rephrased result must maintain the same reasoning process as the original text; 
2. Do not rephrase the following words in the original text, including [catastrophic, catastrophically, bad, badly, excellent, excellently, serious, seriously, poor, poorly, obvious, obviously, fair, fairly, moderate, moderately, good, well]; 
3. Use diverse and natural language. 
The text is: [Desc.]}

\subsubsection{Summarization}
\label{supp:s2i_prompt_s}

\textit{\#User:
You're given a caption of the video and a text on quality assessment that reflects the reasoning process for evaluating video quality. 
Summarize the video caption and the video quality assessment into one complete text written by a quality critic. 
You may refer to the caption of the video as though you are truly seeing this video, but please focus solely on the quality-related content. 
When the caption of the video conflicts with the given video quality assessment, follow the video quality assessment. 
Use diverse and natural language. 
Do not change the following words in the video quality assessment, including [catastrophic, catastrophically, bad, badly, excellent, excellently, serious, seriously, poor, poorly, obvious, obviously, fair, fairly, moderate, moderately, good, well]. 
Do not include the word 'image' in the 
final output. 
Do not imagine and give irrelevant or groundless responses regarding the given video quality assessment. 
The caption of the video is: [Cap.], and the video quality assessment is: [Desc.].
}

\section{Prompt Design in Progressive Tuning}
\label{supp:train}

The prompt design of the tuning in \textit{Score2Instruct} section is as follows.

\subsection{Stage I}
\label{supp:train_1}

\textit{\#User:} \texttt{<img>} \textit{Rate the} \texttt{<dimension>} \textit{of the video.}

\noindent \textit{\#Assistant:} \textit{The} \texttt{<definition>} \textit{is} \texttt{<level>}.

\subsection{Stage II}
\label{supp:train_2}

There are two types of instructions in \textit{Stage II}: quality justifications and question-answering pairs. 
We only need to design prompts for quality justifications.
For \textit{question}, we first prompt \cite{vicuna} to generate 50 candidate questions. 
Subsequently, we manually eliminate ambiguous and repetitive ones and correct inaccurate ones, creating a question set of 20 questions.
Last, we apply these 20 questions to prompt the models on 100 videos. 
By examining the models' responses, we eliminate questions that yielded unsatisfactory results across all models, ultimately refining the selection to 16 questions.
The question pool is as follows: 

\begin{itemize}
    \item \textit{\#User: Provide a brief overview of the video and examine its quality, drawing conclusions from your analysis.}
    \item \textit{\#User: Summarize the video briefly and evaluate its quality features, determining its overall quality based on your observations.}
    \item \textit{\#User: Give a brief description of the video, analyze and evaluate its quality, and draw conclusions from your assessment.}
    \item \textit{\#User: Summarize the video briefly, explore its characteristics, and provide feedback based on your review.}
    \item \textit{\#User: Offer a brief description of the video, closely examine its quality, and present an evaluation based on your analysis.}
    \item \textit{\#User: Briefly describe the video, analyze its quality aspects, and assess it based on your findings.}
    \item \textit{\#User: Provide a brief overview of the video, investigate its quality factors, and present an evaluation based on your insights.}
    \item \textit{\#User: Briefly describe the video, conduct a thorough examination of its quality, and rate it according to your evaluation.}
    \item \textit{\#User: Provide a brief assessment of the video's distortion and visual attributes.}
    \item \textit{\#User: Offer a concise evaluation of the distortion and visual features of the video.}
    \item \textit{\#User: Deliver a short critique of the video's distortion and visual characteristics.}
    \item \textit{\#User: Summarize the distortion and visual attributes of the video in a brief manner.}
    \item \textit{\#User: Give a succinct review of the distortion and visual aspects of the video.}
    \item \textit{\#User: Provide a short analysis of the video's distortion and visual attributes.}
    \item \textit{\#User: Offer a brief overview of the distortion and visual elements present in the video.}
    \item \textit{\#User: Assess the video's distortion and visual attributes in a concise way.}
\end{itemize}

\noindent During training, we randomly pick one question from the question pool.
Here, we omit the video token \texttt{<img>} for readability, the video token is randomly appended to the start or end of the question.

\section{Details of the S2I-Bench}
\label{supp:bench}

\subsection{Open-sourced LLM as Judge}
We provide overall scores using the \textit{open-source} Qwen2.5-32B as the judge. We also conduct a \textit{user study} with 20 participants to measure the interpretability. The results are similar to GPT (Tab.\ref{tab:rank}), proving the metrics' reliability.

\begin{table}[t]
\centering
\caption{Overall score and ranking using different judges, including GPT (adopted), Qwen2.5-32B, and humans.}
\label{tab:rank}
\scalebox{0.56}{
\begin{tabular}{c|cccccc}
\toprule
\textbf{Judge} & LLaVA-OV & LLaVA-Next & InternVL & Video-LLaVA & LLaVA-Video & VideoLLaMA3\\ 
\hline
Qwen & 9.76/rank1 & 9.12/rank4 & 9.68/rank2 & 8.66/rank6 & 9.08/rank5 & 9.36/rank3\\
GPT & 9.88/rank1 & 9.01/rank5 & 9.72/rank2 & 8.50/rank6 & 9.08/rank4 & 9.27/rank3\\
Human & 9.45/rank1 & 9.08/rank5 & 9.38/rank2 & 8.82/rank6 & 9.15/rank4 & 9.30/rank3\\
\bottomrule
\end{tabular}}
\end{table}

\begin{figure}[t]
  \centering
  \includegraphics[width=\linewidth]{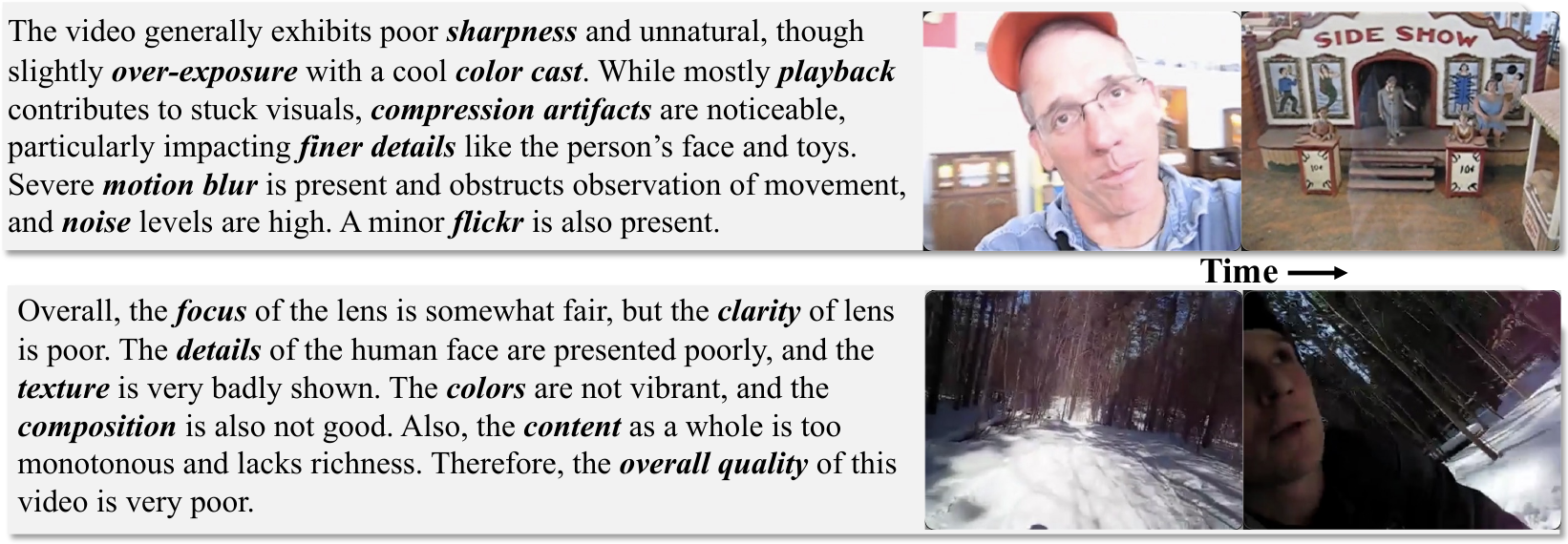}
  \caption{
  More visualized cases of S2I-Bench}
  \label{fig:cases}
\end{figure}
\subsection{Zero-shot Performances of Proprietary Models}
\label{supp:bench_exp}

\begin{table}[t]
\centering
\caption{Evaluations of proprietary models on S2I-Bench.}
\label{tab:LMMs+}
\scalebox{0.8}{
\begin{tabular}{c|cccc|c}
\toprule
\multicolumn{1}{c:}{\textbf{Model}} & \textit{CI}$\uparrow$ & \textit{CU}$\uparrow$ & \textit{DO}$\uparrow$ & \textit{TU}$\uparrow$ & \textbf{\textit{Sum}}$\uparrow$ \\ 
\hline
\multicolumn{1}{c:}{GPT-4o} & 2.28 & 2.23 & 2.11 & 2.22 & \textbf{8.84} \\
\multicolumn{1}{c:}{GPT-4o mini} & 2.11 & 2.08 & 2.09 & 1.97 & 8.25 \\
\multicolumn{1}{c:}{Gemini 1.5 Pro} & 2.21 & 2.22 & 2.15 & 1.95 & 8.53 \\
\hdashline
\multicolumn{1}{c:}{VideoLLaMA3-7B \textit{(baseline)}} & 2.14 & 2.14 & 1.97 & 2.06 & \textbf{8.31} \\
\multicolumn{1}{c:}{LLaVA-Video-7B \textit{(baseline)}} & 2.05 & 2.08 & 2.06 & 1.98 & 8.17 \\
\bottomrule
\end{tabular}}
\end{table}
We further test the zero-shot performances of three closed-source in Tab.\ref{tab:LMMs+}, including GPT-4o \cite{gpt4o}, GPT-4o-mini \cite{gpt4omini}, and Gemini-1.5 Pro \cite{gemini1.5pro}. 
The closed-source models outperform the open-source baseline models in \textit{Tab.2 of the manuscript}. 
Still, the S2I-tuned models in \textit{Tab.2 of the manuscript} remain superior, showing the efficacy of our method.
\subsection{Human Checking in Benchmark Construction}
\label{supp:bench_check}

In \textit{S2I-Bench} section, we uniformly sample 400 video-justification pairs from S2I. 
To ensure the benchmark's reliability, we conduct a thorough manual check as follows:
A total of 20 visual experts conduct thorough filtering and correction to minimise self-evaluation bias of S2I-Bench.
The experts examine 14 quality dimensions of all 400 videos, completing missing dimension ratings, correcting wrong quality ratings and inaccurate high-level content descriptions.
The human checking process is time-consuming, although it is way better than writing ground-truth justifications from scratch. 
Therefore, we opt to scale up the S2I-Bench in the future by checking more videos.
\section{Evaluation Prompt Design}
\label{supp:prompt}
In \textit{Main Results} section, six S2I-tuned video LMMs are evaluated in quality scoring and justification tasks. The evaluation prompts for the two tasks are as follows:
\subsection{Quality Scoring}
\label{supp:prompt_s}

\textit{\#User: Rate the overall quality of the video.}

\noindent \textit{\#Assistant: The overall quality of the video is}
\subsection{Quality Justification}
\label{supp:prompt_j}

\textit{\#User: Briefly describe the video, analyze its quality aspects, and assess it based on your findings.}

\noindent \textit{\#Assistant: }
\subsection{GPT Prompts for VCG Scores}
\label{supp:prompt_vcg}
\subsubsection{Correctness of Information}
\label{supp:prompt_vcg_ci}

\textit{\#System: You are an intelligent chatbot designed for evaluating the factual accuracy of generative outputs for video-based question-answer pairs. Your task is to compare the predicted answer with the correct answer and determine if they are factually consistent. Here's how you can accomplish the task: 
------\#\#INSTRUCTIONS: 
- Focus on the factual consistency between the predicted answer and the correct answer. The predicted answer should not contain any misinterpretations or misinformation.
- The predicted answer must be factually accurate and align with the video content.
- Consider synonyms or paraphrases as valid matches.
- Evaluate the factual accuracy of the prediction compared to the answer.
}

\noindent \textit{\#User: Please evaluate the following video-based question-answer pair:
Question: [question]
f"Correct Answer: [answer]
Predicted Answer: [pred]
Provide your evaluation only as a factual accuracy score where the factual accuracy score is an integer value between 0 and 5, with 5 indicating the highest level of factual consistency.
Please generate the response in the form of a Python dictionary string with keys 'score', where its value is the factual accuracy score in INTEGER, not STRING.
DO NOT PROVIDE ANY OTHER OUTPUT TEXT OR EXPLANATION. Only provide the Python dictionary string.
For example, your response should look like this: {''score': 4.8}.}
\subsubsection{Detail Orientation}
\label{supp:prompt_vcg_do}

\textit{\#System: You are an intelligent chatbot designed for evaluating the detail orientation of generative outputs for video-based question-answer pairs. 
Your task is to compare the predicted answer with the correct answer and determine its level of detail, considering both completeness and specificity. Here's how you can accomplish the task:
------\#\#INSTRUCTIONS: 
- Check if the predicted answer covers all major points from the video. The response should not leave out any key aspects.
- Evaluate whether the predicted answer includes specific details rather than just generic points. It should provide comprehensive information that is tied to specific elements of the video.
- Consider synonyms or paraphrases as valid matches.
- Provide a single evaluation score that reflects the level of detail orientation of the prediction, considering both completeness and specificity.}

\noindent \textit{\#User: Please evaluate the following video-based question-answer pair:
Question: [question]
Correct Answer: [answer]
Predicted Answer: [pred]
Provide your evaluation only as a detail orientation score where the detail orientation score is an integer value between 0 and 5, with 5 indicating the highest level of detail orientation.
Please generate the response in the form of a Python dictionary string with keys 'score', where its value is the detail orientation score in INTEGER, not STRING.
DO NOT PROVIDE ANY OTHER OUTPUT TEXT OR EXPLANATION. Only provide the Python dictionary string.
For example, your response should look like this: {''score': 4.8}.}
\subsubsection{Contextual Understanding}
\label{supp:prompt_vcg_cu}

\textit{\#System: You are an intelligent chatbot designed for evaluating the contextual understanding of generative outputs for video-based question-answer pairs.
Your task is to compare the predicted answer with the correct answer and determine if the generated response aligns with the overall context of the video content. Here's how you can accomplish the task:
------
\#\#INSTRUCTIONS:
- Evaluate whether the predicted answer aligns with the overall context of the video content. It should not provide information that is out of context or misaligned.
- The predicted answer must capture the main themes and sentiments of the video.
- Consider synonyms or paraphrases as valid matches.
- Provide your evaluation of the contextual understanding of the prediction compared to the answer.}

\noindent \textit{\#User: Please evaluate the following video-based question-answer pair:
Question: [question]
Correct Answer: [answer]
Predicted Answer: [pred]
Provide your evaluation only as a contextual understanding score where the contextual understanding score is an integer value between 0 and 5, with 5 indicating the highest level of contextual understanding.
Please generate the response in the form of a Python dictionary string with keys 'score', where its value is contextual understanding score in INTEGER, not STRING.
DO NOT PROVIDE ANY OTHER OUTPUT TEXT OR EXPLANATION. Only provide the Python dictionary string. 
For example, your response should look like this: {''score': 4.8}.}
\subsubsection{Temporal Understanding}
\label{supp:prompt_vcg_tu}

\textit{\#System: You are an intelligent chatbot designed for evaluating the temporal understanding of generative outputs for video-based question-answer pairs.
Your task is to compare the predicted answer with the correct answer and determine if they correctly reflect the temporal sequence of events in the video content. Here's how you can accomplish the task
------
\#\#INSTRUCTIONS: "
- Focus on the temporal consistency between the predicted answer and the correct answer. The predicted answer should correctly reflect the sequence of events or details as they are presented in the video content.
- Consider synonyms or paraphrases as valid matches, but only if the temporal order is maintained.
- Evaluate the temporal accuracy of the prediction compared to the answer.}

\noindent \textit{\#User: Please evaluate the following video-based question-answer pair:
Question: [question]
Correct Answer: [answer]
Predicted Answer: [pred]
Provide your evaluation only as a temporal accuracy score where the temporal accuracy score is an integer value between 0 and 5, with 5 indicating the highest level of temporal consistency.
Please generate the response in the form of a Python dictionary string with keys 'score', where its value is the temporal accuracy score in INTEGER, not STRING.
DO NOT PROVIDE ANY OTHER OUTPUT TEXT OR EXPLANATION. Only provide the Python dictionary string.
For example, your response should look like this: {''score': 4.8}.}

\end{document}